\documentclass[11pt]{article}
\usepackage{acl2002}
\usepackage{times}
\usepackage{latexsym}
\title{A Method for Open-Vocabulary Speech-Driven Text Retrieval}

\author{Atsushi Fujii\thanks{~~~The first and second authors
       are also members of CREST, Japan Science and Technology
       Corporation.}\\
  University of Library and\\
  Information Science\\
  1-2 Kasuga, Tsukuba\\
  305-8550, Japan\\
  {\tt fujii@ulis.ac.jp} \And
  Katunobu Itou\\
  National Institute of\\
  Advanced Industrial\\
  Science and Technology\\
  1-1-1 Chuuou Daini Umezono\\
  Tsukuba, 305-8568, Japan\\
  {\tt itou@ni.aist.go.jp} \And
  Tetsuya Ishikawa\\
  University of Library and\\
  Information Science\\
  1-2 Kasuga, Tsukuba\\
  305-8550, Japan\\
  {\tt ishikawa@ulis.ac.jp}}

\date{}

\newcommand{\etal}{et~al.}
\newcommand{\etaleos}{et~al}
\newcommand{\eq}[1]{(\ref{#1})}

\input{psfig.tex}

\begin{document}
\maketitle
\begin{abstract}
  While recent retrieval techniques do not limit the number of index
  terms, out-of-vocabulary (OOV) words are crucial in speech
  recognition. Aiming at retrieving information with spoken queries,
  we fill the gap between speech recognition and text retrieval in
  terms of the vocabulary size. Given a spoken query, we generate a
  transcription and detect OOV words through speech recognition. We
  then correspond detected OOV words to terms indexed in a target
  collection to complete the transcription, and search the collection
  for documents relevant to the completed transcription. We show the
  effectiveness of our method by way of experiments.
\end{abstract}

\section{Introduction}
\label{sec:introduction}

Automatic speech recognition, which decodes human voice to generate
transcriptions, has of late become a practical technology. It is
feasible that speech recognition is used in real-world human language
applications, such as information retrieval.

Initiated partially by TREC-6, various methods have been proposed for
``spoken document retrieval (SDR),'' in which written queries are used to
search speech archives for relevant
information~\cite{garofolo:trec-97}.  State-of-the-art SDR methods,
where speech recognition error rate is 20-30\%, are comparable with
text retrieval methods in performance~\cite{jourlin:sc-2000}, and thus
are already practical. Possible rationales include that recognition
errors are overshadowed by a large number of words correctly
transcribed in target documents.

However, ``speech-driven retrieval,'' where spoken queries are used to
retrieve (textual) information, has not fully been explored, although
it is related to numerous keyboard-less applications, such as
telephone-based retrieval, car navigation systems, and user-friendly
interfaces.

Unlike spoken document retrieval, speech-driven retrieval is still a
challenging task, because recognition errors in short queries
considerably decrease retrieval accuracy.  A number of references
addressing this issue can be found in past research literature.

Barnett~\etal~\shortcite{barnett:eurospeech-97}  and
Crestani~\shortcite{crestani:fqas-2000} independently performed
comparative experiments related to speech-driven retrieval, where the
DRAGON speech recognition system was used as an input interface for
the INQUERY text retrieval system.  They used as test queries 35
topics in the TREC collection, dictated by a single male speaker.
However, these cases focused on improving text retrieval methods and
did not address problems in improving speech recognition.  As a
result, errors in recognizing spoken queries (error rate was
approximately 30\%) considerably decreased the retrieval accuracy.

Although we showed that the use of target document collections in
producing language models for speech recognition significantly
improved the performance of speech-driven
retrieval~\cite{fujii:springer-2002,itou:asru-2001}, a number of
issues still remain open questions.

Section~\ref{sec:problem} clarifies problems addressed in this paper.
Section~\ref{sec:overview} overviews our speech-driven text retrieval
system. Sections~\ref{sec:speech_recognition}-\ref{sec:query_completion}
elaborate on our methodology.  Section~\ref{sec:experimentation}
describes comparative experiments, in which an existing IR test
collection was used to evaluate the effectiveness of our
method. Section~\ref{sec:related_work} discusses related research
literature.

\section{Problem Statement}
\label{sec:problem}

One major problem in speech-driven retrieval is related to
out-of-vocabulary (OOV) words.

On the one hand, recent IR systems do not limit the vocabulary size
(i.e., the number of index terms), and can be seen as open-vocabulary
systems, which allow users to input any keywords contained in a target
collection.  It is often the case that a couple of million terms are
indexed for a single IR system.

On the other hand, state-of-the-art speech recognition systems still
need to limit the vocabulary size (i.e., the number of words in a
dictionary), due to problems in estimating statistical language
models~\cite{young:ieee-spm-1996} and constraints associated with
hardware, such as memories. In addition, computation time is crucial
for a real-time usage, including speech-driven retrieval.  In view of
these problems,  for many languages the vocabulary size is limited to
a couple of ten
thousands~\cite{itou:jas-1999,paul:darpa-ws-1992,steeneken:eurospeech-1995},
which is incomparably smaller than the size of indexes for practical
IR systems.

In addition, high-frequency words, such as functional words and common
nouns, are usually included in dictionaries and recognized with a high
accuracy. However, those words are not necessarily useful for
retrieval.  On the contrary, low-frequency words appearing in specific
documents are often effective query terms.

To sum up, the OOV problem is inherent in speech-driven retrieval, and
we need to fill the gap between speech recognition and text retrieval
in terms of the vocabulary size. In this paper, we propose a method to
resolve this problem aiming at open-vocabulary speech-driven retrieval.

\section{System Overview}
\label{sec:overview}

\begin{figure}[htbp]
  \begin{center}
    \leavevmode \psfig{file=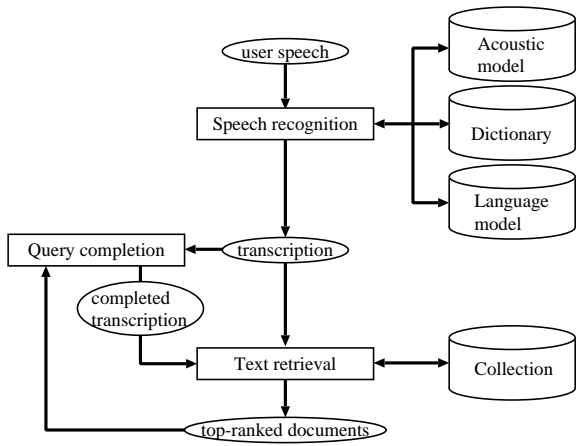,height=2.3in}
  \end{center}
  \caption{The overall design of our speech-driven text retrieval system.}
  \label{fig:system}
\end{figure}

Figure~\ref{fig:system} depicts the overall design of our
speech-driven text retrieval system, which consists of speech
recognition, text retrieval and query completion modules.  Although
our system is currently implemented for Japanese, our methodology is
language-independent.  We explain the retrieval process based on this
figure.

Given a query spoken by a user, the speech recognition module uses a
dictionary and acoustic/language models to generate a transcription of
the user speech. During this process, OOV words, which are not listed
in the dictionary, are also detected.  For this purpose, our language
model includes both words and syllables so that OOV words are
transcribed as sequences of syllables.

For example, in the case where ``{\it kankitsu\/}~(citrus)'' is not
listed in the dictionary, this word should be transcribed as
/\verb|ka N ki tsu|/. However, it is possible that this word is mistakenly
transcribed, such as /\verb|ka N ke tsu|/ and /\verb|ka N ke tsu ke ko|/.

To improve the quality of our system, these syllable sequences have to
be transcribed as {\em words\/}, which is one of the central issues in
this paper. In the case of speech-driven retrieval, where users
usually have specific information needs, it is feasible that users
utter contents related to a target collection. In other words, there
is a great possibility that detected OOV words can be identified as
index terms that are phonetically identical or similar.

However, since a) a single sound can potentially correspond to more
than one word (i.e., homonyms) and b) searching the entire collection
for phonetically identical/similar terms is prohibitive,  we need an
efficient disambiguation method. Specifically, in the case of
Japanese,  the homonym problem is multiply crucial because words
consist of different character types, i.e., ``{\it kanji},'' ``{\it
  katakana},'' ``{\it hiragana},'' alphabets and other characters like
numerals\footnote{In Japanese, {\it kanji\/} (or Chinese character) is
  the idiogram, and {\it katakana\/} and {\it hiragana\/} are
  phonograms.}.

To resolve this problem, we use a two-stage retrieval method. In the
first stage, we delete OOV words from the transcription, and perform
text retrieval using remaining words, to obtain a specific number of
top-ranked documents according to the degree of relevance.  Even if
speech recognition is not perfect, these documents are potentially
associated with the user speech more than the entire collection.
Thus, we search only these documents for index terms corresponding to
detected OOV words.

Then, in the second stage, we replace detected OOV words with
identified index terms so as to complete the transcription, and
re-perform text retrieval to obtain final outputs. However, we do not
re-perform speech recognition in the second stage.

In the above example, let us assume that the user also utters words
related to ``{\it kankitsu\/}~(citrus),'' such as ``{\it
  orenji\/}~(orange)'' and ``{\it remon\/}~(lemon),'' and that these
words are correctly recognized as words. In this case, it is possible
that retrieved documents contain the word ``{\it
  kankitsu\/}~(citrus).'' Thus, we replace the syllable sequence
/\verb|ka N ke tsu|/ in the query with ``{\it kankitsu\/},'' which is
additionally used as a query term in the second stage.

It may be argued that our method resembles the notion of
pseudo-relevance feedback (or local feedback) for IR, where documents
obtained in the first stage are used to expand query terms, and final
outputs are refined in the second stage~\cite{kwok:sigir-98}.
However, while relevance feedback is used to improve only the
retrieval accuracy, our method improves the speech recognition and
retrieval accuracy.

\section{Speech Recognition}
\label{sec:speech_recognition}

The speech recognition module generates word sequence $W$, given phone
sequence $X$.  In a stochastic speech recognition
framework~\cite{bahl:ieee-tpami-1983}, the task is to select the $W$
maximizing $P(W|X)$, which is transformed as in Equation~\eq{eq:bayes}
through the Bayesian theorem.
\begin{equation}
  \label{eq:bayes}
  \arg\max_{W}P(W|X) = \arg\max_{W}P(X|W)\cdot P(W)
\end{equation}
Here, $P(X|W)$ models a probability that word sequence $W$ is
transformed into phone sequence $X$, and $P(W)$ models a probability
that $W$ is linguistically acceptable. These factors are usually
called acoustic and language models, respectively.

For the speech recognition module, we use the Japanese dictation
toolkit~\cite{kawahara:icslp-2000}\footnote{http://winnie.kuis.kyoto-u.ac.jp/dictation},
which includes the ``Julius'' recognition engine and acoustic/language
models. The acoustic model was produced by way of the ASJ speech
database (ASJ-JNAS)~\cite{itou:98:a,itou:jas-1999}, which contains
approximately 20,000 sentences uttered by 132 speakers including the
both gender groups.

This toolkit also includes development softwares so that acoustic and
language models can be produced and replaced depending on the
application.  While we use the acoustic model provided in the toolkit,
we use a new language model including both words and syllables.  For
this purpose, we used the ``ChaSen'' morphological
analyzer\footnote{http://chasen.aist-nara.ac.jp} to extract words from
ten years worth of ``Mainichi Shimbun'' newspaper articles (1991-2000).

Then, we selected 20,000 high-frequency words to produce a dictionary.
At the same time, we segmented remaining lower-frequency words into
syllables based on the Japanese phonogram system. The resultant number
of syllable types was approximately 700.  Finally, we produced a
word/syllable-based trigram language model.  In other words, OOV words
were modeled as sequences of syllables. Thus, by using our language
model, OOV words can easily be detected.

In spoken document retrieval, an open-vocabulary method, which
combines recognition methods for words and syllables in target speech
documents, was also proposed~\cite{wechsler:sigir-98}.  However, this
method requires an additional computation for recognizing syllables,
and thus is expensive. In contrast, since our language model is a
regular statistical $N$-gram model, we can use the same speech
recognition framework as in Equation~\eq{eq:bayes}.

\section{Text Retrieval}
\label{sec:text_retrieval}

The text retrieval module is based on the ``Okapi'' probabilistic
retrieval method~\cite{robertson:sigir-94}, which is used to compute
the relevance score between the transcribed query and each document in
a target collection.  To produce an inverted file (i.e., an index), we
use ChaSen to extract content words from documents as terms, and
perform a word-based indexing. We also extract terms from transcribed
queries using the same method.

\section{Query Completion}
\label{sec:query_completion}

\subsection{Overview}
\label{subsec:qc_overview}

As explained in Section~\ref{sec:overview}, the basis of the query
completion module is to correspond OOV words detected by speech
recognition (Section~\ref{sec:speech_recognition}) to index terms used
for text retrieval (Section~\ref{sec:text_retrieval}). However, to
identify corresponding index terms efficiently, we limit the number of
documents in the first stage retrieval.  In principle, terms that are
indexed in top-ranked documents (those retrieved in the first stage)
and have the same sound with detected OOV words can be corresponding
terms.

However, a single sound often corresponds to multiple words.  In
addition, since speech recognition on a syllable-by-syllable basis is
not perfect, it is possible that OOV words are incorrectly
transcribed.  For example, in some cases the Japanese word ``{\it
  kankitsu\/}~(citrus)'' is transcribed as /\verb|ka N ke tsu|/.
Thus, we also need to consider index terms that are phonetically {\em
  similar\/} to OOV words.  To sum up, we need a disambiguation method
to select appropriate corresponding terms, out of a number of
candidates.

\subsection{Formalization}
\label{subsec:qc_formalization}

Intuitively, it is feasible that appropriate terms:
\begin{itemize}
\item have identical/similar sound with OOV words detected in spoken
  queries,
\item frequently appear in a top-ranked document set,
\item and appear in higher-ranked documents.
\end{itemize}
From the viewpoint of probability theory, possible representations for
the above three properties include Equation~\eq{eq:prob}, where each
property corresponds to different parameters. Our task is to select
the $t$ maximizing the value computed by this equation as the
corresponding term for OOV word $w$.
\begin{equation}
  \label{eq:prob}
  \sum_{d \in D_{q}} P(w|t) \cdot P(t|d) \cdot P(d|q)
\end{equation}
Here, $D_{q}$ is the top-ranked document set retrieved in the first
stage, given query $q$.  \mbox{$P(w|t)$} is a probability that index
term $t$ can be replaced with detected OOV word $w$, in terms of
phonetics. \mbox{$P(t|d)$} is the relative frequency of term $t$ in
document $d$. \mbox{$P(d|q)$} is a probability that document $d$ is
relevant to query $q$, which is associated with the score formalized
in the Okapi method.

However, from the viewpoint of empiricism, Equation~\eq{eq:prob} is
not necessarily effective. First, it is not easy to estimate $P(w|t)$
based on the probability theory. Second, the probability score
computed by the Okapi method is an approximation focused mainly on
{\em relative\/} superiority among retrieved documents, and thus it is
difficult to estimate $P(d|q)$ in a rigorous manner. Finally, it is
also difficult to determine the degree to which each parameter
influences in the final probability score.

In view of these problems, through preliminary experiments we
approximated Equation~\eq{eq:prob} and formalized a method to compute
the degree (not the probability) to which given index term $t$
corresponds to OOV word $w$.

First, we estimate \mbox{$P(w|t)$} by the ratio between the number of
syllables commonly included in both $w$ and $t$ and the total number
of syllables in $w$.  We use a DP matching method to identify the
number of cases related to deletion, insertion, and substitution in
$w$, on a syllable-by-syllable basis.

Second, \mbox{$P(w|t)$} should be more influential than
\mbox{$P(t|d)$} and \mbox{$P(d|q)$} in Equation~\eq{eq:prob}, although
the last two parameters are effective in the case where a large number
of candidates phonetically similar to $w$ are obtained. To decrease
the effect of \mbox{$P(t|d)$} and \mbox{$P(d|q)$}, we tentatively use
logarithms of these parameters. In addition, we use the score computed
by the Okapi method as \mbox{$P(d|q)$}.

According to the above approximation, we compute the score of $t$ as
in Equation~\eq{eq:score}.
\begin{equation}
  \label{eq:score}
  \sum_{d \in D_{q}} P(w|t) \cdot \log(P(t|d) \cdot P(d|q))
\end{equation}
It should be noted that Equation~\eq{eq:score} is independent of the
indexing method used, and therefore $t$ can be any sequences of
characters contained in $D_{q}$. In other words, any types of indexing
methods (e.g., word-based and phrase-based indexing methods) can be
used in our framework.

\subsection{Implementation}
\label{subsec:qc_implementation}

Since computation time is crucial for a real-time usage, we preprocess
documents in a target collection so as to identify candidate terms
efficiently. This process is similar to the indexing process performed
in the text retrieval module.

In the case of text retrieval, index terms are organized in an
inverted file so that documents including terms that {\em exactly\/}
match with query keywords can be retrieved efficiently.

However, in the case of query completion, terms that are included in
top-ranked documents need to be retrieved. In addition, to minimize a
score computation (for example, DP matching is time-consuming), it is
desirable to delete terms that are associated with a diminished
phonetic similarity value, \mbox{$P(w|t)$}, prior to the computation
of Equation~\eq{eq:score}.  In other words, an index file for query
completion has to be organized so that a {\em partial\/} matching
method can be used. For example, /\verb|ka N ki tsu|/ has to be
retrieved efficiently in response to /\verb|ka N ke tsu|/.

Thus, we implemented a forward/backward partial-matching method, in
which entries can be retrieved by any substrings from the first/last
characters. In addition, we index words and word-based bigrams,
because preliminary experiments showed that OOV words detected by our
speech recognition module are usually single words or short phrases,
such as ``{\it ozon-houru\/}~(ozone hole).''

\section{Experimentation}
\label{sec:experimentation}

\subsection{Methodology}
\label{subsec:ex_method}

\begin{figure*}[htbp]
  \begin{center}
    \leavevmode
    \begin{quote}
      \tt
      \footnotesize
      <TOPIC><TOPIC-ID>1001</TOPIC-ID> \\
      <DESCRIPTION>Corporate merging</DESCRIPTION> \\
      <NARRATIVE>The article describes a corporate merging and in the
      article, the name of companies have to be
      identifiable. Information
      including the field and the purpose of the merging have to be
      identifiable. Corporate merging includes corporate acquisition,
      corporate unifications and corporate buying.</NARRATIVE></TOPIC>
    \end{quote}
    \caption{An English translation for an example topic in the IREX collection.}
    \label{fig:irex_topic}
  \end{center}
\end{figure*}

To evaluate the performance of our speech-driven retrieval system, we
used the IREX
collection\footnote{http://cs.nyu.edu/cs/projects/proteus/irex/index-e.html}. This
test collection, which resembles one used in the TREC ad hoc retrieval
track, includes 30 Japanese topics (information need) and relevance
assessment (correct judgement) for each topic, along with target
documents.  The target documents are 211,853 articles collected from
two years worth of ``Mainichi Shimbun'' newspaper (1994-1995).

Each topic consists of the ID, description and narrative. While
descriptions are short phrases related to the topic, narratives
consist of one or more sentences describing the
topic. Figure~\ref{fig:irex_topic} shows an example topic in the SGML
form (translated into English by one of the organizers of the IREX
workshop).

However, since the IREX collection does not contain spoken queries, we
asked four speakers (two males/females) to dictate the narrative
field. Thus, we produced four different sets of 30 spoken queries. By
using those queries, we compared the following different methods:
\begin{enumerate}
\item text-to-text retrieval, which used written narratives as
  queries, and can be seen as a perfect speech-driven text retrieval,
\item speech-driven text retrieval, in which only words listed in the
  dictionary were modeled in the language model (in other words, the
  OOV word detection and query completion modules were not used),
\item speech-driven text retrieval, in which OOV words detected in
  spoken queries were simply deleted (in other words, the query
  completion module was not used),
\item speech-driven text retrieval, in which our method proposed in
  Section~\ref{sec:overview} was used.
\end{enumerate}
In cases of methods~2-4, queries dictated by four speakers were used
independently. Thus, in practice we compared 13 different retrieval
results.  In addition, for methods~2-4, ten years worth of {\it
  Mainichi Shimbun\/} Japanese newspaper articles (1991-2000) were
used to produce language models. However, while method~2 used only
20,000 high-frequency words for language modeling, methods~3 and 4
also used syllables extracted from lower-frequency words (see
Section~\ref{sec:speech_recognition}).

Following the IREX workshop, each method retrieved 300 top documents
in response to each query, and non-interpolated average precision
values were used to evaluate each method.

\subsection{Results}
\label{subsec:ex_results}

First, we evaluated the performance of detecting OOV words. In the 30
queries used for our evaluation, 14 word {\em tokens\/} (13 word {\em
  types\/}) were OOV words unlisted in the dictionary for speech
recognition. Table~\ref{tab:oov_evaluation} shows the results on a
speaker-by-speaker basis, where ``\#Detected'' and ``\#Correct''
denote the total number of OOV words detected by our method and the
number of OOV words correctly detected, respectively.  In addition,
``\#Completed'' denotes the number of detected OOV words that were
corresponded to correct index terms in 300 top documents.

\begin{table*}[htbp]
  \begin{center}
    \caption{Results for detecting and completing OOV words.}
    \medskip
    \leavevmode
    \small
    \begin{tabular}{lcccccc} \hline\hline
      Speaker & \#Detected & \#Correct & \#Completed & Recall &
      Precision & Accuracy \\ \hline
      Female \#1 &  51 &  9 & 18 & 0.643 & 0.176 & 0.353 \\
      Female \#2 &  56 & 10 & 18 & 0.714 & 0.179 & 0.321 \\
      Male \#1   &  33 &  9 & 12 & 0.643 & 0.273 & 0.364 \\
      Male \#2   &  37 & 12 & 16 & 0.857 & 0.324 & 0.432 \\
      \hline
      Total      & 176 & 40 & 64 & 0.714 & 0.226 & 0.362 \\
      \hline
    \end{tabular}
    \label{tab:oov_evaluation}
  \end{center}
\end{table*}

It should be noted that ``\#Completed'' was greater than ``\#Correct''
because our method often mistakenly detected words in the dictionary
as OOV words, but completed them with index terms correctly. We
estimated recall and precision for detecting OOV words, and accuracy
for query completion, as in Equation~\eq{eq:rpa}.
\begin{equation}
  \label{eq:rpa}
  \begin{array}{lll}
    recall & = & \frac{\textstyle \#Correct}{\textstyle 14} \\
    \noalign{\vskip 1.2ex}
    precision & = & \frac{\textstyle \#Correct}{\textstyle \#Detect} \\
    \noalign{\vskip 1.2ex}
    accuracy & = & \frac{\textstyle \#Completed}{\textstyle \#Detect}
  \end{array}
\end{equation}
Looking at Table~\ref{tab:oov_evaluation}, one can see that recall was
generally greater than precision. In other words, our method tended to
detect as many OOV words as possible. In addition, accuracy of
query completion was relatively low.

Figure~\ref{fig:examples} shows example words in spoken queries,
detected as OOV words and correctly completed with index terms. In
this figure, OOV words are transcribed with syllables, where
``\verb|:|'' denotes a long vowel. Hyphens are inserted between
Japanese words, which inherently lack lexical segmentation.

\begin{figure*}[htbp]
  \begin{center}
    \small
    \begin{tabular}{lll} \hline\hline
      {\hfill\centering OOV words\hfill} &
      {\hfill\centering Index terms (syllables)\hfill} &
      {\hfill\centering English gloss\hfill} \\ \hline
      /\verb|gu re : pu ra chi na ga no|/
      & {\it gureepu-furuutsu\/}~/\verb|gu re : pu fu ru : tsu|/
      & grapefruit \\
      /\verb|ya yo i chi ta|/
      & {\it Yayoi-jidai\/}~/\verb|ya yo i ji da i|/
      & the {\it Yayoi\/} period \\
      /\verb|ni ku ku ra i su|/
      & {\it nikku-puraisu\/}~/\verb|ni q ku pu ra i su|/
      & Nick Price \\
      /\verb|be N pi|/
      & {\it benpi\/}~/\verb|be N pi|/
      & constipation \\
      \hline
    \end{tabular}
    \caption{Example words detected as OOV words and completed
      correctly by our method.}
    \label{fig:examples}
  \end{center}
\end{figure*}

Second, to evaluate the effectiveness of our query completion method
more carefully, we compared retrieval accuracy for methods~1-4 (see
Section~\ref{subsec:ex_method}).  Table~\ref{tab:avg_pre} shows
average precision values, averaged over the 30 queries, for each
method\footnote{Average precision is often used to evaluate IR
  systems, which should not be confused with evaluation measures in
  Equation~\eq{eq:rpa}.}. The average precision values of our method
(i.e., method~4) was approximately 87\% of that for text-to-text
retrieval.

By comparing methods~2-4, one can see that our method improved average
precision values of the other methods irrespective of the speaker.  To
put it more precisely, by comparing methods~3 and 4, one can see the
effectiveness of the query completion method. In addition, by
comparing methods~2 and 4, one can see that a combination of the OOV
word detection and query completion methods was effective.

It may be argued that the improvement was relatively small. However,
since the number of OOV words inherent in 30 queries was only 14, the
effect of our method was overshadowed by a large number of other
words. In fact, the number of words used as query terms for our
method, averaged over the four speakers, was 421. Since existing test
collections for IR research were not produced to explore the OOV
problem, it is difficult to derive conclusions that are statistically
valid. Experiments using larger-scale test collections where the OOV
problem is more crucial need to be further explored.

Finally, we investigated the time efficiency of our method, and found
that CPU time required for the query completion process per detected
OOV word was 3.5 seconds (AMD Athlon MP 1900+). However, an additional
CPU time for detecting OOV words, which can be performed in a
conventional speech recognition process, was not crucial.

\subsection{Analyzing Errors}
\label{subsec:error_analysis}

We manually analyzed seven cases where the average precision value of
our method was significantly lower than that obtained with method~2
(the total number of cases was the product of numbers of queries and
speakers).

Among these seven cases, in five cases our query completion method
selected incorrect index terms, although correct index terms were
included in top-ranked documents obtained with the first stage.  For
example, in the case of the query 1021 dictated by a female speaker,
the word ``{\it seido\/}~(institution)'' was mistakenly transcribed as
/\verb|se N do|/. As a result, the word ``{\it sendo\/}~(freshness),''
which is associated with the same syllable sequences, was selected as
the index term. The word ``{\it seido\/}~(institution)'' was the third
candidate based on the score computed by Equation~\eq{eq:score}. To
reduce these errors, we need to enhance the score computation.

In another case, our speech recognition module did not correctly
recognize words in the dictionary, and decreased the retrieval
accuracy.

In the final case, a fragment of a narrative sentence consisting of
ten words was detected as a single OOV word. As a result, our method,
which can complete up to two word sequences, mistakenly processed that
word, and decreased the retrieval accuracy.  However, this case was
exceptional. In most cases, functional words, which were recognized
with a high accuracy, segmented OOV words into shorter fragments.

\begin{table}[htbp]
  \begin{center}
    \caption{Non-interpolated average precision values, averaged over
      30 queries, for different methods.}
    \medskip
    \leavevmode
    \small
    \tabcolsep=4pt
    \begin{tabular}{lcccc} \hline\hline
      Speaker$\backslash$Method & 1 & 2 & 3 & 4 \\ \hline
      Female \#1 & -- & 0.2831 & 0.2834 & 0.3195 \\
      Female \#2 & -- & 0.2745 & 0.2443 & 0.2846 \\
      Male \#1   & -- & 0.3005 & 0.2987 & 0.3179 \\
      Male \#2   & -- & 0.2787 & 0.2675 & 0.2957 \\
      \hline
      Total      & 0.3486 & 0.2842 & 0.2734 & 0.3044 \\
      \hline
    \end{tabular}
    \label{tab:avg_pre}
  \end{center}
\end{table}

\section{Related Work}
\label{sec:related_work}

The method proposed by Kupiec~\etal~\shortcite{kupiec:arpa-hlt-94} and
our method are similar in the sense that both methods use target
collections as language models for speech recognition to realize
open-vocabulary speech-driven retrieval.

Kupiec~\etaleos's method, which is based on word recognition and
accepts only short queries, derives multiple transcription candidates
(i.e., possible word combinations), and searches a target collection
for the most plausible word combination. However, in the case of
longer queries, the number of candidates increases, and thus the
searching cost is prohibitive.  This is a reason why operational
speech recognition systems have to limit the vocabulary size.

In contrast, our method, which is based on a recent {\em continuous\/}
speech recognition framework, can accept longer
sentences. Additionally, our method uses a two-stage retrieval
principle to limit a search space in a target collection, and
disambiguates only detected OOV words. Thus, the computation cost can
be minimized.

\section{Conclusion}
\label{sec:conclusion}

To facilitate retrieving information by spoken queries, the
out-of-vocabulary problem in speech recognition needs to be
resolved. In our proposed method, out-of-vocabulary words in a query
are detected by speech recognition, and completed with terms indexed
for text retrieval, so as to improve the recognition accuracy. In
addition, the completed query is used to improve the retrieval
accuracy. We showed the effectiveness of our method by using dictated
queries in the IREX collection.  Future work would include experiments
using larger-scale test collections in various domains.

\bibliographystyle{acl.bst}

\end{document}